\pdfoutput=1

\documentclass[11pt]{article}

\usepackage[preprint]{acl}

\usepackage{times}
\usepackage{latexsym}

\usepackage[T1]{fontenc}

\usepackage[utf8]{inputenc}

\usepackage{microtype}
\usepackage{dblfloatfix}
\usepackage{graphicx}
\usepackage{amsmath}
\usepackage{amssymb}
\usepackage{booktabs}
\usepackage{multirow}
\usepackage{multicol}
\usepackage{adjustbox}
\usepackage{subcaption}
\usepackage[linesnumbered,ruled]{algorithm2e}
\usepackage[noend]{algpseudocode}
\usepackage{breqn}
\SetKwInput{KwInput}{Input}                
\SetKwInput{KwOutput}{Output}              
\usepackage{inconsolata}

\usepackage{graphicx}
\newcommand*{\aide}{\textsc{Aide}}

\newcount\Comments  
\usepackage{color}
\newcommand{\kibitz}[2]{\ifnum\Comments=1\textcolor{#1}{#2}\fi}

\title{\aide: Agentically Improve Visual Language Model with Domain Experts}

\author{
 \textbf{Ming-Chang Chiu\textsuperscript{1}},
 \textbf{Fuxiao Liu\textsuperscript{2}},
 \textbf{Karan Sapra\textsuperscript{3}},
 \textbf{Andrew Tao\textsuperscript{3}},
\\
 \textbf{Yaser Yacoob\textsuperscript{2}},
 \textbf{Xuezhe Ma\textsuperscript{1}},
 \textbf{Zhiding Yu\textsuperscript{3}},
 \textbf{Guilin Liu \textsuperscript{3}},
\\
 \textsuperscript{1}University of Southern California,
 \textsuperscript{2}University of Maryland, College Park,
 \textsuperscript{3}NVIDIA,
}

\begin{document}
\maketitle
\begin{abstract}
The enhancement of Visual Language Models (VLMs) has traditionally relied on knowledge distillation from larger, more capable models. This dependence creates a fundamental bottleneck for improving state-of-the-art systems, particularly when no superior models exist. We introduce \aide~(Agentic Improvement through Domain Experts), a novel framework that enables VLMs to autonomously enhance their capabilities by leveraging specialized domain expert models. \aide~operates through a four-stage process: (1) identifying instances for refinement, (2) engaging domain experts for targeted analysis, (3) synthesizing expert outputs with existing data, and (4) integrating enhanced instances into the training pipeline. Experiments on multiple benchmarks, including MMMU, MME, MMBench, etc., demonstrate \aide's ability to achieve notable performance gains without relying on larger VLMs nor human supervision. Our framework provides a scalable, resource-efficient approach to continuous VLM improvement, addressing critical limitations in current methodologies, particularly valuable when larger models are unavailable to access.
\end{abstract}

\section{Introduction}
\vspace{-1mm}
Visual Language Models (VLMs) have achieved impressive advancements in understanding and reasoning about visual content \citep{jeanbaptiste2022flamingo,haotian2023visual,yunhao2024vila2}. However, their continued improvement often hinges on knowledge distillation from larger, more capable models through approaches like instruction tuning \citep{fuxiao2023mitigating,haotian2023visual,liu2023mmc}. While this approach has proven effective for intermediate-scale models, it introduces a significant limitation for the largest state-of-the-art systems: the absence of a superior model renders further enhancement infeasible. This "chicken-and-egg" problem stifles progress and raises a critical question: how can VLMs be improved when no superior models exist? Despite their general capabilities, VLMs often underperform in specialized tasks compared to domain expert models such as object segmentation tools or Optical Character Recognition (OCR) systems. For instance, models like Grounding DINO \citep{shilong2023grounding} consistently outperform general-purpose VLMs \citep{lu2021florence,shaohan2023language} in visual recognition tasks (Table~\ref{tab:refcomp}). This observation suggests an alternative pathway: rather than relying on larger general models, VLMs can leverage the specialized capabilities of expert models for improvement \cite{shi2024eagle}.

\begin{figure*}[tbp]
    \centering
    \includegraphics[width=\linewidth]{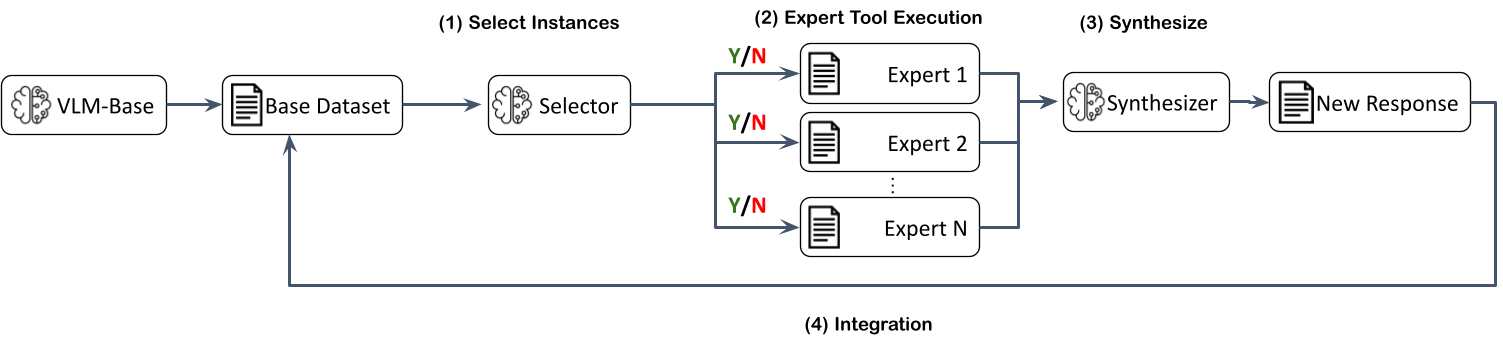}\vspace{-2mm}
    \caption{\textbf{\aide~Workflow}. \aide consists of two agents, a Selector and a Synthesizer. The Selector interacts with the data instances and autonomously invoke the expert tools as it deems fit. The Synthesizer collects information from the original data instances along with outputs from the select experts and generate enriched response.}
    \label{fig:workflow}
\end{figure*}

In this paper, we introduce \aide~(Agentic Improvement through Domain Experts), a framework that enables VLMs to strategically collaborate with domain expert models to enhance their training data. As shown in Fig. \ref{fig:workflow}. \aide~employs a four-stage workflow: (1) identifying instances requiring refinement, (2) invoking expert models for specialized outputs, (3) synthesizing these outputs with existing data, and (4) systematically integrating improved data points into the training process.

We validate \aide~'s effectiveness through extensive experiments on benchmarks such as MMMU \citep{yue2024mmmu}, MME \citep{fu2024mmecomprehensiveevaluationbenchmark}, MMBench \citep{liu2024mmbench}, etc., showing that it achieves notable performance improvements using only off-the-shelf lightweight expert models. Unlike traditional methods, \aide~does not depend on access to larger models nor human supervision, making it a scalable and computationally efficient solution for advancing state-of-the-art VLMs.

\begin{table*}[ht]

\centering
\begin{adjustbox}{width=0.7\linewidth}
\begin{tabular}{l|ccc|ccc|cc}
\toprule
\textbf{Model} &  \multicolumn{3}{c}{RefCOCO} & \multicolumn{3}{c}{RefCOCO+} & \multicolumn{2}{c}{RefCOCOg} \\ 
                &    val & testA & testB & val & testA & testB & val & test \\ \hline
 \multicolumn{9}{c}{general-purpose model} \\\hline
Kosmos-2   & 52.3 & 57.4 & 47.3 & 45.5 & 50.7 & 42.2 & 60.6 & 61.7 \\ 
Florence-2-B & 53.9 & 58.4 & 49.7 & 51.5 & 56.4 & 47.9 & 66.3 & 65.1 \\ 
Florence-2-L & 56.3 & 61.6 & 51.4 & 53.6 & 57.9 & 49.9 & 68.0 & 67.0 \\ \hline
 \multicolumn{9}{c}{expert model} \\\midrule
Grounding DINO L & \textbf{90.56} & \textbf{93.19} & \textbf{88.24} & \textbf{82.75} & \textbf{88.95} & \textbf{75.92} & \textbf{86.13} & \textbf{87.02} \\ 
\bottomrule
\end{tabular}
\end{adjustbox}
\caption{Performance comparison between general-purpose models and expert models on referring expression comprehension tasks. Expert model like Grounding DINO outperforms general-purpose models by a large margin.\vspace{-2mm}}
\label{tab:refcomp}
\end{table*}
\section{Related Work}

\paragraph{Knowledge Distillation.} Traditional methods for improving VLMs rely on knowledge distillation \citep{wang2022self}, where a larger ``teacher" model generates training data to enhance a smaller ``student" model. While effective for intermediate-scale models \citep{haotian2023visual, fuxiao2023mitigating}, this paradigm creates a dependency on the availability of superior models, which limits its applicability to state-of-the-art systems. 

\paragraph{Specialized Models.}Recent studies \cite{fei2024multimodal} highlight the superiority of domain-specific expert models in certain tasks. For example, object detection systems such as Grounding DINO and OCR models like PaddleOCR \cite{paddleocr} significantly outperform general-purpose VLMs in their respective domains \citep{lu2021florence, tianhe2024grounded, shilong2023grounding}. These findings underscore the potential of leveraging specialized models to complement the general capabilities of VLMs.
\vspace{-2mm}
\paragraph{Data Synthesis and Augmentation.} Existing methods for augmenting training data often involve the model generating synthetic examples \citep{fuxiao2023mitigating, chen2023sharegpt4v} or applying templates to initial human annotations for more truthful data \citep{chiu2024megacoinenhancingmediumgrainedcolor, chiu2024colorsensestudycolorvision}. While this approach can enhance performance on specific benchmarks, it risks perpetuating the biases and limitations of the model, resulting in diminishing returns. In contrast, \aide~integrates external expert knowledge and the original samples into the data generation pipeline, enabling more robust and unbiased improvements.

\vspace{-2mm}
\section{\aide~Framework}
\vspace{-2mm}
The \aide~framework enables VLMs to autonomously improve by collaborating with domain expert models. It comprises two primary agents—Selector and Synthesizer—and operates through three principal actions: Selection, Execution, and Synthesis. \aide presumes an existing base dataset as the environment for agents to interact with. Fig.~\ref{fig:workflow} provides an overview of the \aide~pipeline.

\vspace{-3mm}
\paragraph{Selector.}
The selector serves two objectives, identify improvement candidates and match candidates with expert tools: the selector interacts with the base dataset and is presented with detailed information and functionalities of the expert tools and judge if any of the additional information the experts can provide may be beneficial to improve the quality of the data. If it is, then the selector will exercise the corresponding expert tool.


\begin{table*}[t]

\centering
\begin{adjustbox}{max width=\textwidth}
\begin{tabular}{lllllllllllllll}
\toprule
\textbf{Selector} & \textbf{ChartQA}  & \textbf{TextVQA} & \textbf{Mathvista} & \textbf{MMbench} & \textbf{MME-c} & \textbf{MME-p} & \textbf{MMMU\_val} & \textbf{SciQA} & \textbf{POPE} \\
\midrule
Eagle-Baseline (reproduced) & 80.24 & 7.469 & 55.8 & 74.1 & 337.5 & 1529.1 & 41.0 & 80.42 & 89.29 \\
\midrule
Eagle Heuristic (ans $\leq$5 tokens) & 79.64  & 41.49 &54.5	&71.0&	336.4	&1511.9	&40.9 &77.05 &89.56\\
Eagle-Small-step (\aide) & 79.64 & \textbf{76.85} & 56.6 & 74.4 & \textbf{385.7} & \textbf{1590.1} & 41.4 & \textbf{82.99} & 89.19 \\
Eagle-Retention (\aide) &  \textbf{80.42} & 74.46 & \textbf{56.9} & \textbf{74.8} & 335.7 & 1583.6 & \textbf{42.2} & 78.83 & \textbf{90.07} \\
\bottomrule
\end{tabular}
\end{adjustbox}
\caption{\textbf{\aide~performance Improvements}. Using VLM for self improvement enhanced the performance on various benchmarks.\vspace{-3mm}}
\label{tab:metrics}
\end{table*}

\begin{figure*}[t]
    \centering
    \includegraphics[width=\linewidth]{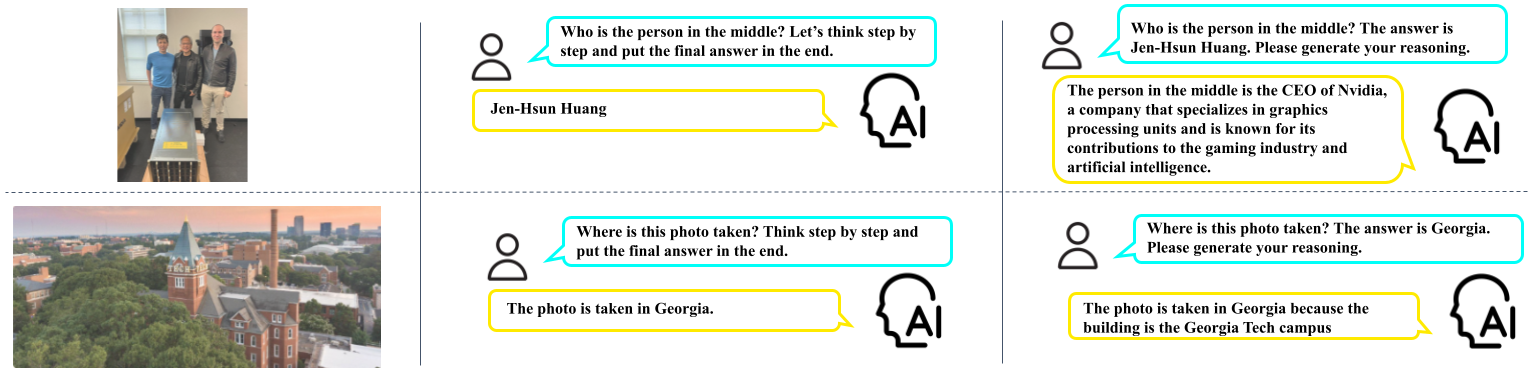}\vspace{-3mm}
    \caption{\textbf{Small-step Prompting}. We observe even when VLM is able to answer the query (\textit{middle-column}), sometimes the instruction following is not stable. And simplifying the prompt into smaller steps by giving the answer (\textit{last column}) gives more detailed responses.}
    \label{fig:prompt_results_illustration}
\end{figure*}

\paragraph{Synthesizer.}
The Synthesizer integrates expert outputs with the original data to generate enhanced training examples. This process involves: \textit{(i) Aggregating information from multiple sources, i.e. the original instances and domain expert outputs; (ii) Resolving potential conflicts. E.g. between original instances and expert outputs. (iii) Producing richer and more coherent responses.} By these instructions, we expect the new response s would inherently be richer and contain reasoning flavors. See Sec.~\ref{sec:qualitative} for comparisons.

\vspace{-3mm}
\subsection{Integration}
\vspace{-1mm}
After generating enhanced samples, the integration incorporates them back into the training pipeline. This involves filtering: ensures new formulations maintain sensible information along with the original instances to prevent model collapse.

\vspace{-1mm}
\section{Experiment}

\vspace{-3mm}
\paragraph{Setup.} We evaluate \aide~using the Eagle-8B \citep{shi2024eagle} as both Selector and Synthesizer, interacting with the Cambrian1-7M dataset for one iteration. Experiments are conducted on an NVIDIA A100 node with 8 GPUs. Note that the choice of Selector and Synthesizer can be adaptable and need not be the same.

\vspace{-3mm}
\paragraph{Expert Tool Choice.} Two lightweight domain experts, PaddleOCR \citep{paddleocr} and Grounded-SAM \citep{tianhe2024grounded}, are employed. These tools complement the visual data-rich composition of Cambrian1-7M \citep{shengbang2024cambrian1}. \aide~is extensible to incorporate additional expert models for multimodal tasks.

\begin{figure*}
    \centering
    \includegraphics[width=\linewidth]{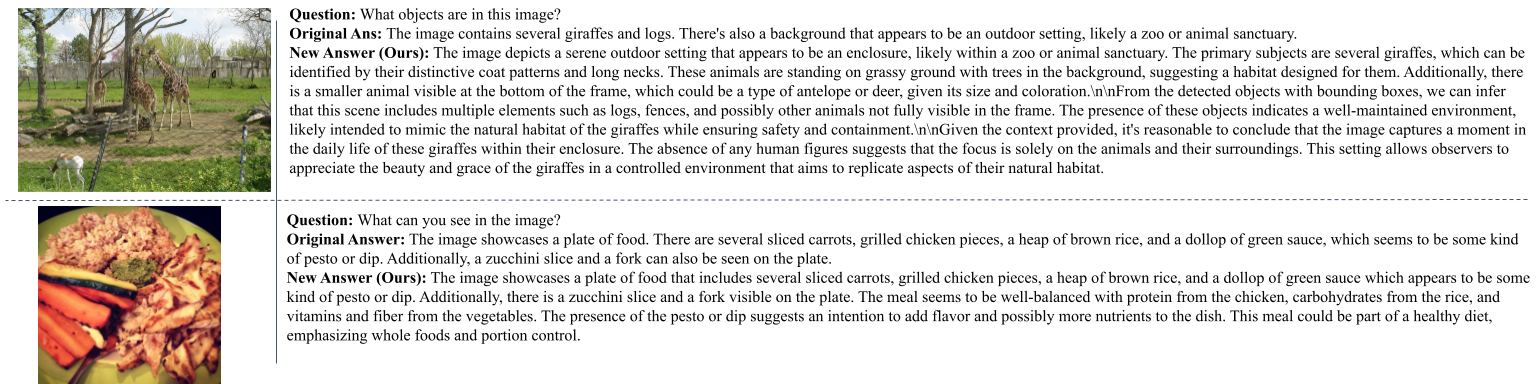}\vspace{-2mm}
    \caption{Comparisons of the original and the new answer produced by \aide. Our \aide~workflow enriches the responses.}\vspace{-5mm}
    \label{fig:qualitative}
\end{figure*}

\vspace{-3mm}
\paragraph{Integration.} We use simple heuristics like n-gram filtering, etc. because we use small-step prompting that we deem enough to maintain quality of new responses (Fig.~\ref{fig:prompt_results_illustration} \& Fig.~\ref{fig:qualitative}), but \aide~can easily add verifiers to further enhance data generation quality (see Sec.~\ref{sec:discussion} for discussions).

\vspace{-3mm}
\paragraph{Results.}

Table~\ref{tab:metrics} shows that applying \aide~is able to improve on MMMU \cite{yue2024mmmu} by 1.2\%, MMBench \cite{MMBench} by 0.77\%, MME \cite{fu2023mme} by 52, Mathvista \cite{lu2024mathvista} by 1.1 \%, ChartQA \cite{masry2022chartqa} by 1.1 \% etc. These results \cite{saikh2022scienceqa, li2023evaluating} highlight \aide~'s effectiveness by using domain expertise for VLM improvement.

\vspace{-3mm}
\subsection{Ablations}
\vspace{-1mm}
\paragraph{Selector Choice.} We evaluated variations in Selector strategies, including text-only LLMs and heuristic methods. Tab.~\ref{tab:metrics} (row 2) shows that heuristic like synthesizing the instances that has $\leq$ 5 tokens is not comparable to using a VLM-selector, even though the heuristic would select much more instances for synthesis (2.5M vs 950k).

\vspace{-3mm}
\paragraph{Small-step Prompting.} In the synthesis step, we tried to directly prompt the VLM to generate more detailed responses and then put final answer at the end, but it often fails to do so (Fig.~\ref{fig:prompt_results_illustration}-\textit{mid}). Fig.~\ref{fig:prompt_results_illustration}-\textit{last} shows that it is effective to simply prompt the VLM with the whole information of the instance and with one task (e.g. just generate the reasoning). Even though the VLM knows the correct answers, slightly more complex prompt cannot achieve the desirable outcome.

\vspace{-3mm}
\paragraph{Originals Retention.} Tab.~\ref{tab:metrics} shows that \aide~is able to provide improvements with or without the original turn (question, answer pair), suggesting the effectiveness of \aide.
\begin{figure*}
    \begin{subfigure}[b]{0.35\textwidth}
         \centering
         \includegraphics[width=\linewidth]{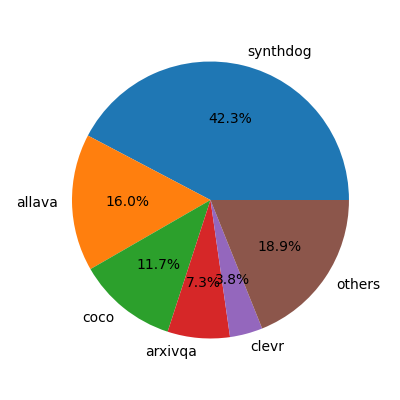}
    
     \end{subfigure}\hfill
    \begin{subfigure}[b]{0.64\textwidth}
         \centering
         \includegraphics[width=\linewidth]{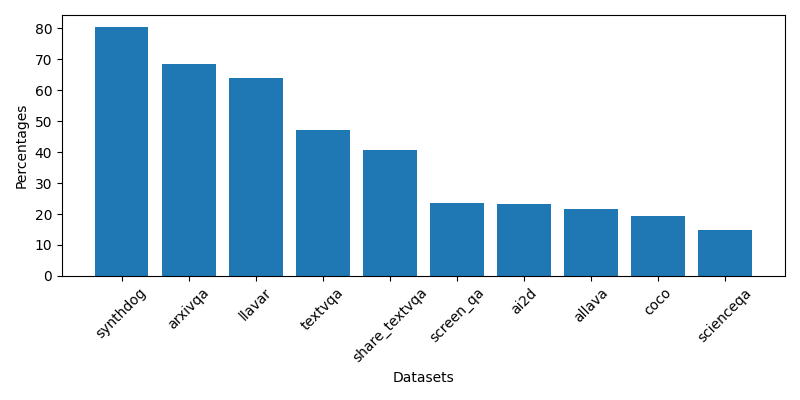}
    
     \end{subfigure}\hfill
     \vspace{-4mm}
    \caption{\textit{Left:} Breakdown of selected data instances by VLM-Selector. Synthdog takes the most proportion of the selection. \textit{Right:} Ratio of data instances selected by the VLM-Selector to the total instances in the original Cambrian-1. }
    \label{fig:distribution}
\end{figure*}

\subsection{Analysis on \aide-selected data} 

We analyze the VLM selected data points for improvement. Out of the 7M training instance, about 2M are text-only and 5M are multimodal. And of the multimodal training instances, around 950K were selected by VLM-Selector. We provide the breakdown of the 950K in Fig.~\ref{fig:distribution}.  
Detailed analysis of \aide-selected data reveals interesting patterns in two directions, the proportions chosen among the 950k and proportions among the multimodal samples in Cambrian1. We observe that the majority (over 40\%) of the selected are from synthdog, an OCR dataset, suggesting the Selector deems the quality of synthdog need most improvement. On the other hand, we analyze the percentage of the selected candidates compared to the original data by source from Cambrian1. Again, about 80\% of the synthdog are chosen by the VLM selector for improvement. Arxivqa llavar, textvqa are also predominately selected. We suspect the Selector deems the quality of these document datasets need improvement and posit \aide~may serve as an alternative way to estimate the quality of a dataset through a VLM-as-a-judge approach.

\subsection{Qualitative results}\label{sec:qualitative}
Figure~\ref{fig:qualitative} illustrates the comparisons between the original data instances and the enriched data instances by our \aide~workflow. The new responses provides more details and reasoning-flavored context than the original answers. These contextual enhancements \citep{chiu2024megacoinenhancingmediumgrainedcolor} may explain \aide's ability to improve the performances on various benchmarks (Tab.~\ref{tab:metrics}).

\vspace{-2mm}
\section{Conclusions}\label{sec:discussion}
\vspace{2mm}
We presented \aide, an agentic framework enabling VLM improvement through domain expert models. Unlike traditional methods, \aide~offers a scalable, resource-efficient alternative to reliance on larger models. Our contributions include: \textit{(i) A novel approach to VLM enhancement without superior models. (ii) Demonstrated improvements across benchmarks like MMMU, MMBench, and SciQA. (iii) Detailed analysis of data selection strategies and their impacts.}

Future work may explore adapting \aide~for preference optimization, generating new (question, answer) pairs and incorporating test-time inference techniques to further guarantee the quality of new data. These advancements aim to further refine synthesized data quality and broaden \aide's applicability, paving the way for continuous VLM training paradigms.

\section{Limitations}\label{sec:Limitations} 
Although \aide~avoids the need for larger VLMs, integrating expert models and synthesizing enhanced data require additional computational resources. While we utilized lightweight models in our experiments, applying \aide~to large-scale datasets or in real-time settings may be constrained by computational costs. Optimizing selection heuristics and developing more efficient integration strategies could enhance scalability.

\bibliography{custom}



\end{document}